\documentclass[11pt,a4paper]{article}
\usepackage[hyperref]{naaclhlt2018}
\usepackage{times}
\usepackage{latexsym}
\usepackage{adjustbox}
\usepackage{amsmath}
\usepackage{hyperref}
\usepackage{array}
\usepackage{ragged2e}
\usepackage{url}
\usepackage{color}
\usepackage{comment}

\usepackage{caption}
\DeclareCaptionFont{10pt}{\fontsize{10pt}{12pt}\selectfont}
\newcolumntype{P}[1]{>{\RaggedRight\hspace{0pt}}p{#1}}
\definecolor{darkgreen}{RGB}{60,90,35}

\newcommand{\yejin}[1]{{\color{cyan}yc:[#1]}}

\aclfinalcopy
\title{Deep Communicating Agents for Abstractive Summarization}
\author{Asli Celikyilmaz$^1$, Antoine Bosselut$^2\thanks{\hspace{2mm}Work done while author was at Microsoft Research}$, Xiaodong He$^3$ \and Yejin Choi$^{2,4}$\\
$^1$Microsoft Research \\
$^2$Paul G. Allen School of Computer Science \& Engineering, University of Washington \\
$^3$JD AI Research \\
$^4$Allen Institute for Artificial Intelligence\\
\texttt{\{aslicel\}@microsoft.com} \: \: \: \texttt{\{xiaodong.he\}@jd.com}\\
\texttt{\{antoineb, yejin\}@cs.washington.edu}
}
\date{}
\begin{document}
\maketitle
\begin{abstract}
We present \textit{deep communicating agents} in an encoder-decoder architecture to address the challenges of representing a long document for abstractive summarization. 
With deep communicating agents, the task of encoding a long text is divided across multiple collaborating agents,
each in charge of a subsection of the input text. These encoders are connected to a single decoder, trained end-to-end using
reinforcement learning to generate a focused and coherent summary. 
Empirical results demonstrate that multiple communicating encoders lead to a higher quality summary compared to several strong baselines, including those based on a single encoder or multiple non-communicating encoders.

\end{abstract}

\section{Introduction}
\label{sec:intro}
We focus on the task of \emph{abstractive} summarization of a \emph{long} document. In contrast to \emph{extractive} summarization, where a summary is composed of a subset of sentences or words lifted from the input text as is, \emph{abstractive} summarization requires the generative ability to rephrase and restructure sentences to compose a coherent and concise summary.
%
As recurrent neural networks (RNNs) are capable of generating fluent language, variants of encoder-decoder RNNs \cite{seq2seq,bahdanau2014neural} have shown promising results on the abstractive summarization task \cite{deepsummrush,summs2s}. 

The fundamental challenge, however, is that the strong performance of neural models at encoding short text does not generalize well to long text. The motivation behind our approach is to be able to dynamically attend to different parts of the input to capture salient facts. 
While recent work in summarization addresses these issues using improved attention models \cite{deepsummchopra}, pointer networks with coverage mechanisms~\cite{summpoinernet}, and coherence-focused training objectives \cite{rlsummsocher,seqtutor}, 
an effective mechanism for representing a long document remains a challenge.

\begin{figure}[t]
\begin{center} 
\adjustbox{trim={0.04\width} {0.35\height} {0.04\width} {0.04\height},clip}%
{
\includegraphics[width=0.50\textwidth]{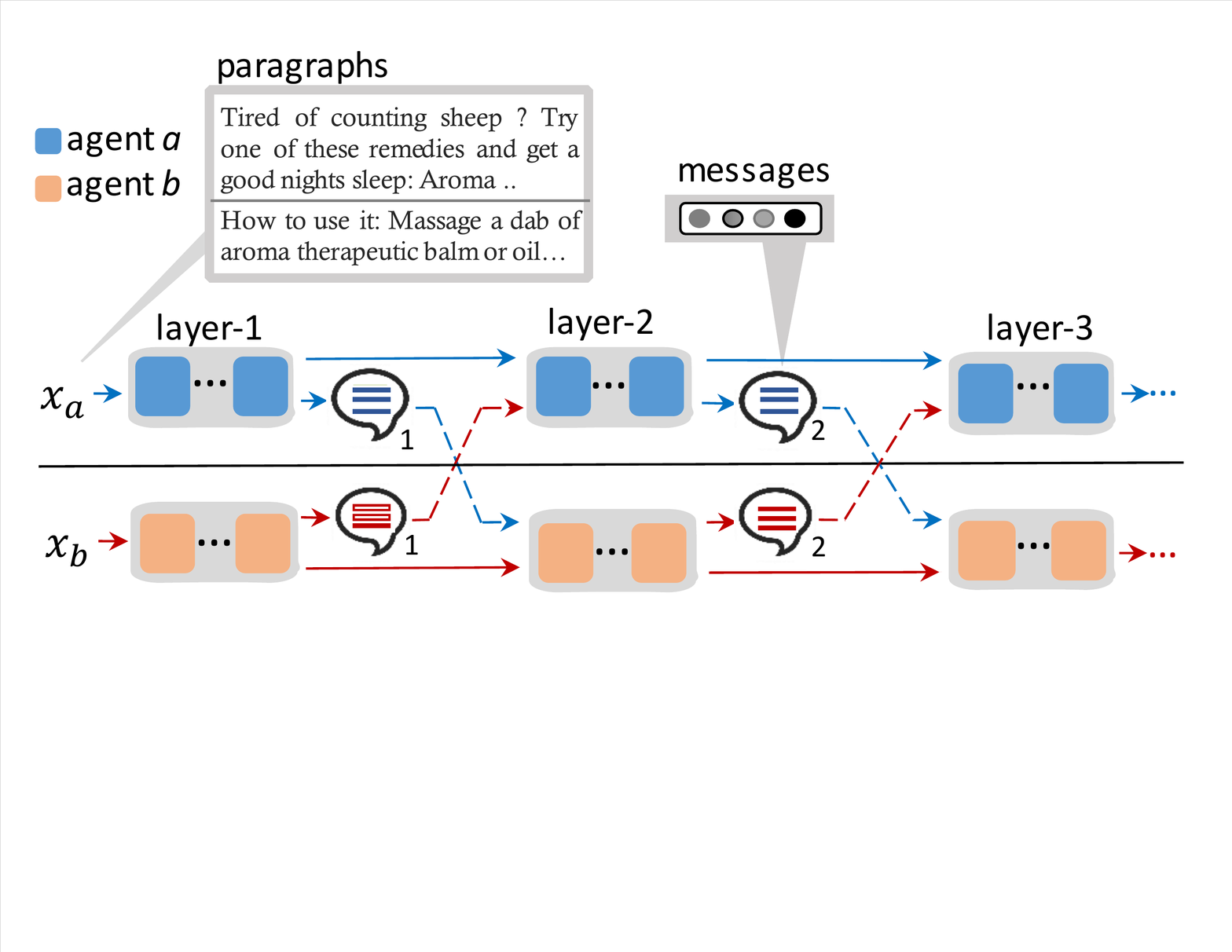}
}
\end{center} 
\vskip -0.27in
\caption{Illustration of deep communicating agents presented in this paper. Each agent $a$ and $b$ encodes one paragraph in multiple layers. 
By passing new messages through multiple layers the agents are able to coordinate and
focus on the important aspects of the input text.}
\label{agentcomm}
\end{figure}

\begin{figure*}[t!]
\begin{center} 
\adjustbox{trim={0.039\width} {0.25\height} {0.038\width} {0.03\height},clip}%
{
\includegraphics[width=0.95\textwidth]{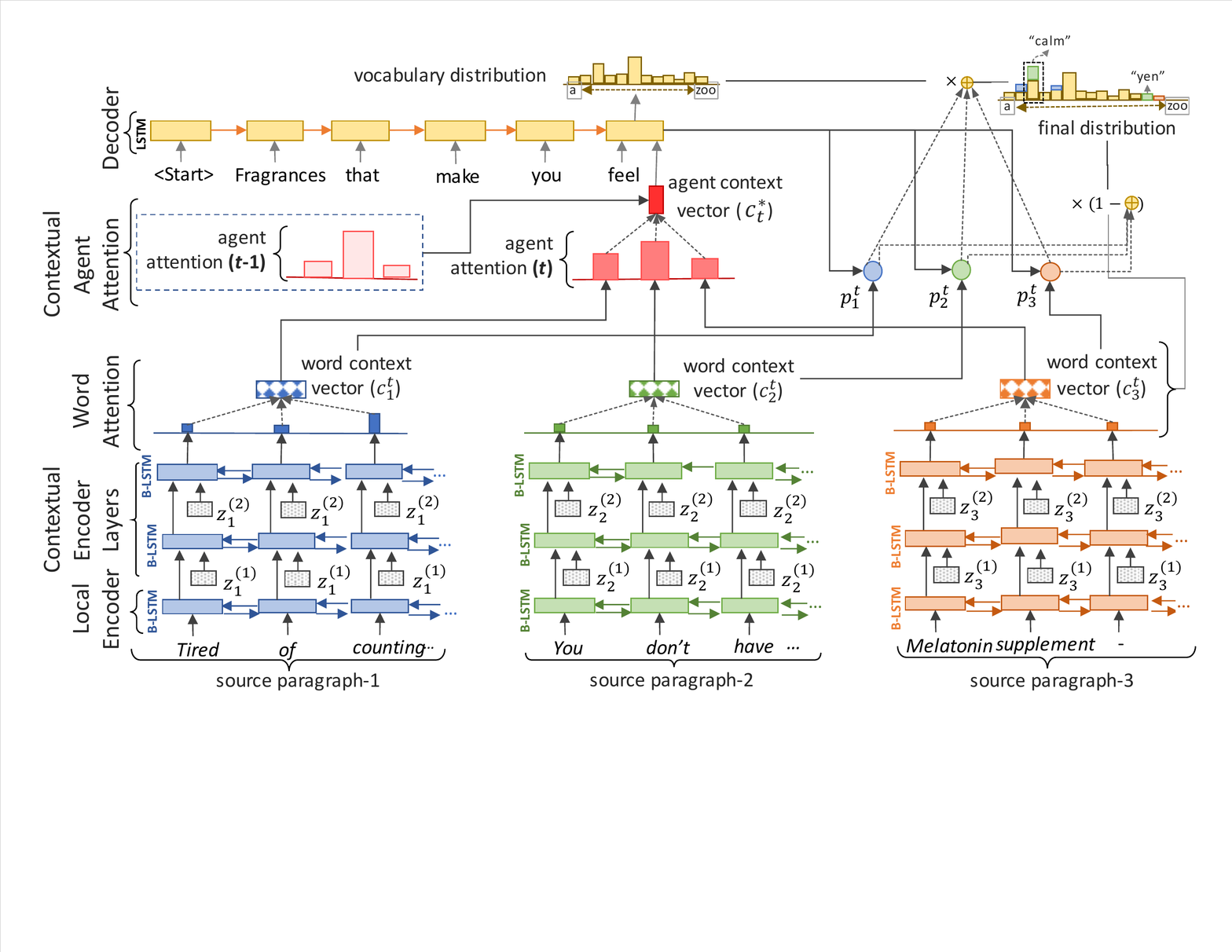}
}
\end{center} 
\vskip -0.25in
\caption{Multi-agent-encoder-decoder overview. Each agent $a$ encodes a paragraph using a local encoder followed by multiple contextual layers with agent communication through concentrated messages $z_a^{(k)}$ at each layer $k$. \textbf{Communication is illustrated in Figure~\ref{fpicmessagepassing}}.
The word context vectors $c^t_a$ are condensed into agent context $c_t^*$. Agent specific generation probabilities, \textit{p}$_a^t$, enable voting for the suitable out-of-vocabulary words (e.g., '\textit{yen}') in the final distribution.}
\label{fpic}
\end{figure*}

Simultaneous work has investigated the use of deep communicating agents \cite{commnet} for collaborative tasks such as logic puzzles \cite{multiagent4}, visual dialog \cite{visualdialog}, and reference games \cite{multiagent2}. Our work builds on these approaches to propose the first study on using communicating agents to encode long text for summarization.

The key idea of our model is to divide the hard task of encoding a long text across multiple collaborating encoder agents, each in charge of a different subsection of the text (Figure~\ref{agentcomm}).
Each of these agents encodes their assigned text independently, and broadcasts their encoding to others, allowing agents to share global context information with one another about different sections of the document. 
All agents then adapt the encoding of their assigned text in light of the global context and repeat the process across multiple layers, generating new messages at each layer.
Once each agent completes encoding, they deliver their information to the decoder with a novel \textit{contextual agent attention} (Figure~\ref{fpic}). Contextual agent attention enables the decoder to integrate information from multiple agents smoothly at each decoding step.  
The network is trained end-to-end using
self-critical reinforcement learning~\cite{scic} to generate focused and coherent summaries. 

Empirical results on the CNN/DailyMail and New York Times datasets demonstrate that multiple communicating encoders lead to higher quality summaries compared to strong baselines, including those based on a single encoder or multiple non-communicating encoders. 
Human evaluations indicate that our model is able to produce more focused summaries. The agents gather salient information from multiple areas of the document, and communicate their information with one another, 
thus reducing common mistakes such as missing key facts, repeating the same content, or including unnecessary details. Further analysis reveals that our model attains better performance when the decoder interacts with multiple agents in a more balanced way, confirming the benefit of representing a long document with multiple encoding agents.  



\section{Model}
We extend the CommNet model of \citet{commnet} for sequence generation. 
\paragraph{Notation}
Each document $d$ is a sequence of paragraphs $x_a$, which are split across multiple encoding agents $a$=1,..,$M$ (e.g., agent-1 encodes the first paragraph $x_1$, agent-2 the second paragraph $x_2$, so on). Each paragraph $x_{a}$=$\{w_{a,i}\}^I$, is a sequence of $I$ words. 
We construct a $V$-sized vocabulary from the training documents from the most frequently appearing words.
Each word $w_{a,i}$ is embedded into a $n$-dimensional vector $\mathbf{e}_{{a,i}}$. 
 All $W$ variables are linear projection matrices.
\subsection{Multi-Agent Encoder}
\label{ucbir}
Each agent 
encodes the word sequences with the following two stacked encoders. 

\noindent{\textbf{Local Encoder}} The first layer is a local encoder of each agent \textit{a}, where 
the tokens of the corresponding paragraph $x_a$ are fed into a single layer bi-directional LSTM (\texttt{bLSTM}), producing the local encoder hidden states, $h_{i}^{(1)} \in \mathcal{R}^{H}$: 
\begin{align}
\overrightarrow{h}_{i}^{(1)},\overleftarrow{h}_{i}^{(1)} = \texttt{bLSTM}(e_{i},\overrightarrow{h}_{i-1}^{(1)},\overleftarrow{h}_{{i}+1}^{(1)})\\
h_{i}^{(1)}= W_1[\overrightarrow{h}_{i}^{(1)},\overleftarrow{h}_{i}^{(1)}] 
\end{align}
where $H$ is the hidden state dimensionality. 
The output of the local encoder layer is fed into the contextual encoder. 

\begin{figure}[t]
\begin{center} 
\adjustbox{trim={0.04\width} {0.3\height} {0.05\width} {0.03\height},clip}%
{
\includegraphics[width=0.45\textwidth]{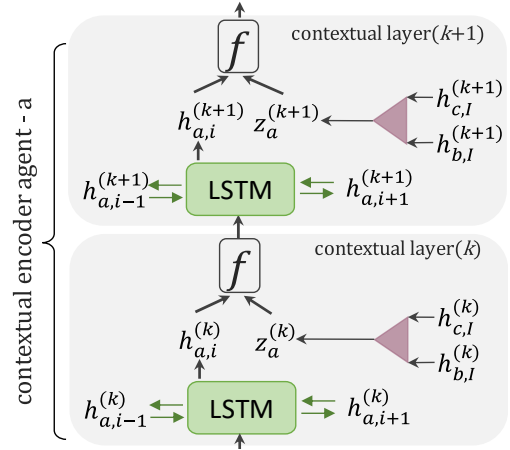}
}
\end{center} 
\vskip -0.19in
\caption{Multi-agent encoder message passing. Agents $b$ and $c$ transmit the last hidden state output ($I$) of the current layer $k$ as a message, which are passed through an average pool (Eq.~\eqref{fmessage}). The receiving agent $a$ uses the new message $\textstyle z_a^{(k)}$ as additional input to its next layer.}
\label{fpicmessagepassing}
\end{figure}

\noindent{\textbf{Contextual Encoder}}
Our framework enables agent communication cycles across multiple encoding layers. 
The output of each contextual encoder is an adapted representation of the agent's encoded information conditioned on the information received from the other agents.
At each layer $k$=1,..,$K$, each agent $a$ jointly encodes the information received from the previous layer (see Figure~\ref{fpicmessagepassing}). 
Each cell of the ($k$+1)th contextual layer is a \texttt{bLSTM} that takes three inputs: the hidden states from the adjacent LSTM cells, $\overrightarrow{h}_{i-1}^{(k+1)}$$\in$$R^{H}$ or $\overleftarrow{h}_{i+1}^{(k+1)}$$\in$$R^{H}$, the hidden state from the previous layer $h_{i}^{(k)}$, and the message vector from other agents $z^{(k)}$$\in$$R^{H}$ and outputs $h_{i}^{(k+1)}$$\in$$R^{H}$:
\begin{align}
\overrightarrow{h}_{i}^{(k+1)},\overleftarrow{h}_{i}^{(k+1)} = &\texttt{bLSTM}(f(h_{i}^{(k)}, z^{(k)}), \\
&\overrightarrow{h}_{i-1}^{(k+1)},\overleftarrow{h}_{{i}+1}^{(k+1)})\\
h_{i}^{(k+1)}&= W_2 [\overrightarrow{h}_{i}^{(k+1)},\overleftarrow{h}_{i}^{(k+1)}] \label{eq:ctx_output}
\end{align}

\noindent where $i$=$1..I$ indicates the index of each token in the sequence. 

The message $z^{(k)}$ received by any agent $a$ in layer $k$ is the average of the outputs of the other agents from layer $k$:

\begin{equation}
z^{(k)}=\textstyle{\frac{1}{M-1}\sum_{m\neq a}} h_{m,I}^{(k)}
\label{fmessage}
\end{equation}
where $h_{m,I}^{(k)}$ is the last hidden state output from the $k$th contextual layer of each agent where $m \neq a$. Here, we take the average of the messages received from other encoder agents, but a parametric function such as a feed forward model or an attention over messages could also be used. 

The message $z^{(k)}$ is projected with the agent's previous encoding of its document:
\begin{equation}
f(h^{(k)}_{i},z^{(k)}) = v_1^{\mathtt{T}}\text{tanh}(W_3h_{i}^{(k)}+W_4z^{(k)})
\label{f}
\end{equation}
where $v_1, W_3$, and $W_4$ are learned parameters shared by every agent. 
Equation~\eqref{f} combines the information sent by other agents with the context of the current token from this paragraph. 
This yields different features about the current context in relation to other topics in the source document. At each layer, the agent modifies its representation of its own context relative to the information received from other agents, and updates the information it sends to other agents accordingly.

\subsection{Decoder with Agent Attention}
\label{uciki}
The output from the last contextual encoder layer of each agent \{$h^{(K)}_{a,i}$\}$^I$, which is a sequence of hidden state vectors of each token $i$, is sent to the decoder to calculate word-attention distributions. 
We use a single-layer LSTM for the decoder and feed the last hidden state from the first agent $s_0 = h^{(K)}_{1,I}$ as the initial state. 
At each time step $t$, the decoder predicts a new word in the summary $w_t$ and computes a new state $s_t$ by attending to relevant input context provided by the agents.

The decoder uses a new hierarchical attention mechanism over the agents. First, 
a \textbf{\textit{word attention}} distribution $l_{a}^t$ (\citet{bahdanau2014neural}) is computed over every token $\{h^{(K)}_{a,i}\}^I$ for each agent $a$:
\begin{align}
l_{a}^t=\textup{\small softmax}(v_2^{\mathtt{T}}\tanh(W_5h^{(K)}_{a} + W_{6}s_t + b_{1}))
\label{attention2}
\end{align}
where $l_{a}^t$$\in$$[0,1]^I$ is the attention over all tokens in a paragraph $x_a$ and $v_2$, $W_5$, $W_{6}$, $b_{1}$ are learned parameters.
For each decoding step $t$, a new decoder context is calculated for each agent:
\begin{align}
c_{a}^t=\sum_i l_{a,i}^th_{a,i}^{(K)}
\label{context}
\end{align}
which is the weighted sum of the encoder hidden states of agent $a$. 
Each \emph{word context vector} represents the information extracted by the agent from the paragraph it has read. 
Here the decoder has to decide which information is more relevant to the current decoding step $t$. This is done by weighting each context vector by an agent attention yielding the document global \textbf{\textit{agent attention}} distribution $g^t$ (see Figure~\ref{fpic}):
\begin{align}
g^t=\textup{\small softmax} (v_3^{\mathtt{T}}\textup{tanh}(W_{7} c^t + W_{8}s_t + b_{2}))
\label{aattention2}
\end{align}
where 
$v_3$, $W_{7}$, $W_{8}$, and $b_{2}$ are learned, and $g^t\in$[0,1]$^M$ is a soft selection over $M$ agents. 
Then, we compute the agent context vector $c^*_t$:
\begin{align}
c^*_t=\textstyle{\sum_a} g_a^tc_{a}^t
\label{context}
\end{align}

\noindent The agent context $c^*_t$$\in$$R^H$ is a fixed length vector encoding salient information from the entire document provided by the agents. 
It is then concatenated with the decoder state $s_t$ and fed through a multi-layer perception to produce a vocabulary distribution (over all vocabulary words) at time $t$:
\begin{equation}
P^{voc}(w_t \vert s_t, w_{t-1}) =\textup{\small softmax} (\textup{MLP}([s_t,c^*_t]))
\label{vocab}
\end{equation}
To keep the topics of generated sentences intact, it is reasonable that the decoder utilize the same agents over the course of short sequences (e.g., within a sentence).  
Because the decoder is designed to select which agent to attend to at each time step, we introduce contextual agent attention (\textit{caa}) to prevent it from frequently switching between agents. 
The previous step's agent attention $c^*_{t-1}$ is used as additional information to the decoding step to generate a distribution over words:
\begin{equation}
P^{voc}(w_t \vert \cdot) =\textup{\small softmax} (\textup{MLP}([s_t,c^*_t,c^*_{t-1}]))
\label{vocabcaa}
\end{equation}

\subsection{Multi-Agent Pointer Network}
\label{sec:mapointernet}

Similar to \citet{summpoinernet}, we allow for copying candidate words from different paragraphs of the document by computing a \textit{generation} probability value for each agent $p_a^t$ $\in$[0,1] at each timestep $t$ using the context vector $c_{a}^t$, decoder state $s_t$ and decoder input $y_t$:
\vspace*{-1mm}
\begin{equation}
p_a^t=\sigma(v_5^{\mathtt{T}}c_{a}^t + v_6^{\mathtt{T}}s_t+v_7^{\mathtt{T}}y_t+b)
\label{pgenagent1}
\end{equation}
where $b$ is a learned scalar, $y_t$ is the ground-truth/predicted output (depending on the training/testing time). The generation probability determines whether to generate a word from the vocabulary by sampling from $P^{voc}(w\vert \cdot)$, or copying a word from the corresponding agent's input paragraph $x_a$ by sampling from its attention distribution $l_a^t$. 
This produces an extended vocabulary that includes words in the document that are considered out-of-vocabulary (OOV).
A probability distribution over the extended vocabulary is computed for each agent:
\begin{equation}
P^a(w_t \vert \cdot )=p_a^tP^{voc}(w_t \vert \cdot) + (1-p_a^t) \textstyle u_{a,w}^t
\label{pgenagent2}
\end{equation}
 where $u_{a,w}^t$ is the sum of the attention for all instances where $w$ appears in the source document.
The final distribution over the extended vocabulary, from which we sample, is obtained by weighting each agent by their corresponding agent attention values $g_a^t$:
\begin{equation}
P(w_t \vert s_t, w_{t-1})=\textstyle \sum_{a} g^t_aP^a(w_t \vert \cdot )
\label{pgenagent3}
\end{equation}

\noindent In contrast to a single-agent baseline \cite{summpoinernet}, our model allows each agent to vote for different OOV words at time $t$ (Equation~\eqref{pgenagent3}). In such a case, only the word that is relevant to the generated summary up to time $t$ is collaboratively voted as a result of agent attention probability $g_a^t$.    
\section{Mixed Objective Learning}
\label{ssec:rewards}

To train the deep communicating agents, we use a mixed training objective that jointly optimizes multiple losses, which we describe below.

\paragraph{MLE}
Our baseline multi-agent model uses maximum likelihood training for sequence generation. Given $y^*=\{y_1^*$,$y_2^*$,...,$y_T^*\}$ as the ground-truth output sequence (human summary word sequences) for a given input document $d$, we minimize the negative log-likelihood of the target word sequence:
\begin{equation}
L_{\text{MLE}}=-\textstyle \sum_{t=1}^N\text{log} p(y_t^*|y_1^*\dots y_{t-1}^*, d)
\label{lossmle}
\end{equation}
\paragraph{Semantic Cohesion}
%
To encourage sentences in the summary to be informative without repetition, 
we include a \emph{semantic cohesion} loss to integrate sentence-level semantics into the learning objective. 
As the decoder 
generates the output word sequence \{$y_1,y_2\dots y_{T}$\}, it keeps track of the end of sentence delimiter token (`.') indices. 
The hidden state vectors at the end of each sentence $s^{\prime}_{q}$, $q$=1$\dots Q$, where $s^{\prime}_{q}$$\in$\{$s_t$:$y_t$=`$\cdot$', 1$\leq$$t$$\leq$$T$\}, are used to compute the cosine similarity between two consecutively generated sentences. 
To minimize the similarity between end-of-sentence hidden states 
we define a \emph{semantic cohesion} loss: 
\begin{equation}
L_{\text{SEM}} =\textstyle \sum_{q=2}^{Q} \cos (s^{\prime}_{q}, s^{\prime}_{q-1}) 
\end{equation}
The final training objective is then:
\vspace*{-1mm}
\begin{equation}
L_{\text{MLE-SEM}} = L_{\text{MLE}} +\lambda L_{\text{SEM}} 
\end{equation}
where $\lambda$ is a tunable hyperparameter. 
\paragraph{Reinforcement Learning (RL) Loss}
Policy gradient methods can directly optimize discrete target evaluation metrics such as ROUGE that are non-differentiable \cite{rlsummsocher,seqtutor,scvc,googlemt}. At each time step, the word generated by the model can be viewed as an action taken by an RL agent.
Once the full sequence $\hat{y}$ is generated, it is compared against the ground truth sequence $y^*$ to compute the reward $r(\hat{y})$. 

Our model learns using a \textit{self-critical training} approach \cite{scic}, which learns by exploring new sequences and comparing them to the best greedily decoded sequence. 
For each training example $d$, two output sequences are generated: $\hat{y}$, which is sampled from the probability distribution at each time step, $p(\hat{y}_t|\hat{y}_1\dots \hat{y}_{t-1}, d)$, and $\tilde{y}$, the baseline output, which is greedily generated by argmax decoding from $p(\tilde{y}_t|\tilde{y}_1\dots \tilde{y}_{t-1}, d)$. 
The training objective is then to minimize: 
\begin{equation}
L_{\text{RL}}=\textstyle (r(\tilde{y})-r(\hat{y})) \sum_{t=1}^N \text{log} p(\hat{y}_t|\hat{y}_1\dots \hat{y}_{t-1}, d)
\label{lossrl}
\end{equation}
This loss ensures that, with better exploration, the model learns to generate sequences  $\hat{y}$ that receive higher rewards compared to the baseline $\tilde{y}$, increasing overall reward expectation of the model. 
\paragraph{Mixed Loss} 
While training with only MLE loss will learn a better language model, this may not guarantee better results on global performance measures. Similarly, optimizing with only RL loss may increase the reward gathered at the expense of diminished readability and fluency of the generated summary \cite{rlsummsocher}. A combination of the two objectives can yield improved task-specific scores while maintaining fluency:
\begin{equation}
L_{\text{MIXED}}= \gamma L_{\text{RL}}+(1-\gamma)L_{\text{MLE}}
\label{lossmixed}
\end{equation}
where $\gamma$ is a tunable hyperparameter used to balance the two objective functions. We pre-train our models with MLE loss, and then switch to the mixed loss.
We can also add the semantic cohesion loss term: $L_{\text{MIXED-SEM}} = \gamma L_{\text{RL}}+(1-\gamma)L_{\text{MLE-SEM}}$ to analyze its impact in RL training.
\paragraph{Intermediate Rewards}
We introduce sentence-based rewards as opposed to end of summary rewards, using differential ROUGE metrics, to promote generating diverse sentences. Rather than rewarding sentences based on the scores obtained at the end of the generated summary, we compute incremental rouge scores of a generated sentence $\hat{o}_q$:
\vspace*{-5mm}
\begin{equation}
r(\hat{o}_q)=r([\hat{o}_{1},\dots \hat{o}_{q}])-r([\hat{o}_{1},\dots \hat{o}_{q-1}])
\end{equation}
Sentences are rewarded for the increase in ROUGE they contribute to the full sequence, ensuring that the current sentence contributed novel information to the overall summary.

\section{Experimental Setup}
\label{sec:experiments}
\noindent{\textbf{Datasets}}
\label{ssec:datasets}
We conducted experiments on two summarization datasets: CNN/DailyMail \cite{summs2s,teachingmachines} and 
New York Times (NYT) \cite{nytdataset}. 
We replicate the preprocessing steps of \citet{rlsummsocher} to obtain the same data splits, except that we do not anonymize named entities. 
For our DCA models, we initialize the number of agents before training, 
and partition the document among the agents (i.e., three agent $\rightarrow$ three paragraphs). 
Additional details can be found in Appendix~\ref{app:datasets}.
\paragraph{Training Details}
During training and testing we truncate the
article to 800 tokens and limit the length of the
summary to 100 tokens for training and 110 tokens
at test time. We distribute the truncated articles among agents for multi-agent models, preserving the paragraph and sentences as possible. 
%
For both datasets, we limit the input and output vocabulary size to the 50,000 most frequent tokens in the training set. 
We train with up to two contextual layers in all the DCA models as more layers did not provide additional performance gains.
We fix $\gamma=0.97$ for the RL term in Equation~\eqref{lossmixed} and $\lambda=0.1$ for the SEM term in MLE and MIXED training. Additional details are provided in Appendix~\ref{app:training}.
\begin{table*}[ht]
	\small
    \centering
    \begin{tabular}{|l | c | c | c |}
         \hline
         Model & ROUGE-1 & ROUGE-2 & ROUGE-L \\
         \hline
         SummaRuNNer \cite{summs2s} & 39.60 &  16.20 & 35.30 \\
         graph-based attention \cite{graphbased} & 38.01 & 13.90 & 34.00 \\
         pointer generator \cite{summpoinernet} & 36.44 & 15.66 & 33.42 \\
         pointer generator + coverage \cite{summpoinernet} & 39.53  & 17.28 & 36.38 \\
         controlled summarization with fixed values \cite{controllable} & 39.75 & 17.29 & 36.54 \\
         RL, with intra-attention \cite{rlsummsocher} & 41.16 & 15.75 & \textbf{39.08} \\
         ML+RL, with intra-attention\cite{rlsummsocher} & 39.87 & 15.82 & 36.90 \\
         \hline
         (\textbf{m1}) MLE, pgen, no-comm (1-agent) (our baseline-1) & 36.12& 14.38& 33.83\\
         (\textbf{m2}) MLE+SEM, pgen, no-comm (1-agent) (our baseline-2) & 36.90 & 15.02& 33.00 \\
         (\textbf{m3}) MLE+RL, pgen, no-comm (1-agent) (our baseline-3) & 38.01& 16.43& 35.49\\
         \hline
         (\textbf{m4}) DCA MLE+SEM, pgen, no-comm (3-agents) & 37.45& 15.90& 34.56\\
         (\textbf{m5}) DCA MLE+SEM, \textit{mpgen}, with-comm (3-agents) & 39.52 & 17.12 & 36.90 \\
         (\textbf{m6}) DCA MLE+SEM, \textit{mpgen}, with-comm, with \textit{caa} (3-agents) & 41.11  & \textbf{18.21}& 36.03\\
         (\textbf{m7}) DCA MLE+SEM+RL, \textit{mpgen}, with-comm, with \textit{caa} (3-agents) & \textbf{41.69} & \textbf{19.47} & 37.92 \\     
         \hline
    \end{tabular}
    \vspace{-0.05in}
    \caption{Comparison results on the \textbf{CNN/Daily Mail} test set using the \textbf{F1} variants of \textbf{Rouge}. Best model models are bolded.}
    \label{tab:summ1}
\end{table*}
\begin{table*}[ht]
	\small
    \centering
    \begin{tabular}{|l | c | c | c |}
         \hline
         Model & Rouge-1 & Rouge-2 & Rouge-L \\
         \hline
         ML, no intra-attention \cite{rlsummsocher} & 44.26 & 27.43 & 40.41 \\        
         RL, no intra-attention \cite{rlsummsocher} & 47.22 & 30.51 & \textbf{43.27}\\
         ML+RL, no intra-attention\cite{rlsummsocher} & 47.03 & 30.72 & 43.10 \\
         \hline         
         (\textbf{m1}) MLE, pgen, no-comm (1-agent) (our baseline-1) & 44.28 & 26.01& 37.87\\
         (\textbf{m2}) MLE+SEM, pgen, no-comm (1-agent) (our baseline-2) &44.50  & 28.04& 38.80\\
         (\textbf{m3}) MLE+RL, pgen, no-comm (1-agent) (our baseline-3) & 46.15&29.50& 39.38\\
         \hline
         (\textbf{m4}) DCA MLE+SEM, pgen, no-comm (3-agents) & 45.84	& 28.23	& 39.32 \\
         (\textbf{m5}) DCA MLE+SEM, \textit{mpgen}, with-comm (3-agents) & 46.20& 30.01& 40.65\\
         (\textbf{m6}) DCA MLE+SEM, \textit{mpgen}, with-comm, with \textit{caa} (3-agents) & \textbf{47.30} & 30.50 & 41.06 \\
         (\textbf{m7}) DCA MLE+SEM+RL, \textit{mpgen} with-comm, with \textit{caa} (3-agents) & \textbf{48.08} & \textbf{31.19} & 42.33\\         
         \hline
    \end{tabular}
    \vspace{-0.05in}
    \caption{Comparison results on the \textbf{New York Times} test set using the \textbf{F1} variants of \textbf{Rouge}. Best model models are bolded.}
    \label{tab:summ2}
\end{table*}
\paragraph{Evaluation}
We evaluate our system using 
ROUGE-1 (unigram recall), ROUGE-2 (bigram recall) and ROUGE-L (longest common sequence).\footnote{We use \texttt{pyrouge} (pypi.python.org/pypi/pyrouge/0.1.3).} 
We select the MLE models with the lowest negative log-likelihood and the MLE+RL models with the highest ROUGE-L scores on a sample of validation data to evaluate on the test set. At test time, we use beam search of width 5 on all our models to generate final predictions.

\noindent{\textbf{Baselines}} 
We compare our DCA models against previously published models: SummaRuNNer \cite{summs2s}, a graph-based attentional neural model \cite{graphbased} an RNN-based extractive summarizer that combines abstractive features during training; Pointer-networks with and without coverage~\cite{summpoinernet}, RL-based training for summarization with intra-decoder attention~\cite{rlsummsocher}),
and Controllable Abstractive Summarization~\cite{controllable} which allows users to define attributes of generated summaries and also uses a copy mechanism for source entities and decoder attention to reduce repetition.

\noindent{\textbf{Ablations}} 
We investigate each new component of our model with a different ablation, producing seven different models. Our first three ablations are: a single-agent model with the same local encoder, context encoder, and pointer network architectures as the DCA encoders trained with MLE loss  (\textbf{m1}); the same model trained with additional semantic cohesion SEM loss (\textbf{m2}), and the same model as the (\textbf{m1}) but trained with a mixed loss and end-of-summary rewards (\textbf{m3}).

The rest of our models use 3 agents and incrementally add one component. First, we add the semantic cohesion loss (\textbf{m4}). Then, we add multi-agent pointer networks (\textit{mpgen}) and agent communication (\textbf{m5}). Finally, we add contextual agent attention (caa) (\textbf{m6}), and train with the mixed MLE+RL+SEM loss (\textbf{m7}). All DCA models use pointer networks.
\section{Results}
\subsection{Quantitative Analysis}
We show our results on the CNN/DailyMail and NYT datasets in Table~\ref{tab:summ1} and ~\ref{tab:summ2} respectively. Overall, our (\textbf{m6}) and (\textbf{m7}) models with multi-agent encoders, pointer generation, and communication are the strongest models on ROUGE-1 and ROUGE-2. While weaker on ROUGE-L than the RL model from \citet{rlsummsocher}, the human evaluations in that work showed that their model received lower \textit{readability} and \textit{relevance} scores than a model trained with MLE, indicating the additional boost in ROUGE-L was not correlated with summary quality. This result can also account for our best models being more abstractive. Our models use mixed loss not just to optimize for sentence level structure similarity with the reference summary (to get higher ROUGE as reward), but also to learn parameters to improve semantic coherence, promoting higher abstraction (see Table~\ref{tab:summ3} and Appendix B 
for generated summary examples).

\begin{table}[h]
	\small
    \centering
\begin{tabular}{lccc}
\hline
Model & ROUGE-1 & ROUGE-2 & ROUGE-L \\ \hline
\hline
2-agent & 40.94 & 19.16 & 37.54 \\
3-agent & \textbf{41.69} & \textbf{19.47} & 37.92 \\
5-agent &  40.99 & 19.02 & \textbf{38.21} \\
\hline
\end{tabular}
\vskip -0.1in
\caption{Comparison of multi-agent models varying the number of agents using ROUGE results of model (m7) from Table~\ref{tab:summ1} on CNN/Daily Maily Dataset.}
\label{multiagentcomparison}
\end{table}
\noindent \textbf{Single vs. Multi-Agents} All multi-agent models show improvements over the single agent baselines.
On the CNN/DailyMail dataset, compared to MLE published baselines, we improve across all ROUGE scores. We found that the 3-agent models generally outperformed both 2- and 5-agent models (see Table~\ref{multiagentcomparison}). 
This 
is in part because 
we truncate documents before training and 
the larger number of agents might be more efficient for multi-document summarization.
\begin{table*}[t]
	\small
    \centering
\begin{center}
    \begin{tabular}{P{1.1cm}@{}| p{14.4cm}}
    \hline
    \small \textbf{Human}  & \small Mr Turnbull was interviewed about his childhood and his political stance. He also admitted he \textbf{{planned to run for prime minister if Tony Abbott had been successfully toppled in February's leadership spill.}} The words 'primed minister' were controversially also printed on the cover.
 \\ \hline
    \textbf{Single} & \small Malcolm Turnbull is set to feature on the front cover of the GQ Australia in a bold move that will no doubt set senators' tongues wagging. \colorbox{pink}{Posing in a suave blue suit with a pinstriped shirt and a contrasting red tie}, Mr Turnbull's confident demeanour is complimented by the bold, confronting words printed across the page: 'primed minister'.
  \\\hline
   \textbf{Multi} & \small Malcolm Turnbull was set to \textbf{run for prime minister if Tony Abbott had been successfully toppled in February's leadership spill}. He is set to feature on the front cover of the liberal party's newsletter.
 \\ \hline
    \end{tabular}
    \begin{tabular}{P{1.1cm}@{}| p{14.4cm}}
    \hline
    \small \textbf{Human}  & \small Daphne Selfe has been modelling since the fifties. She has recently landed a new campaign with vans and \& other stories. \textbf{The 86-year-old commands 1,000 a day for her work.}
 \\ \hline
    \textbf{Single} & \small Daphne Selfe, 86, shows off the collaboration between the footwearsuper-brandand theetherealhigh street store with uncompromisinggrace. Daphne said of the collection , in which she appears with \colorbox{pink}{22-year-old flo dron: 'the} \colorbox{pink}{\& other stories collection that is featured in this story is truly relaxed and timeless with a modern twist'. The} \colorbox{pink}{shoes are then worn with pieces from the brands ss2015 collection.}
  \\\hline
   \textbf{Multi} & \small Daphne Selfe, 86, has starred in the campaign for vans and \& other stories. The model appears with \colorbox{pink}{22-year-old} \colorbox{pink}{flo dron \& other hair collection}. \textbf{She was still commanding 1,000 a day for her work.} 
 \\ \hline
    \end{tabular}
\end{center}
\vspace{-0.2in}
\caption{Comparison of a human summary to best single- and multi-agent model summaries, (m3) and (m7) from CNN/DailyMail dataset. Although single-agent model generates a coherent summary, it is less focused and contains more unnecessary details (\colorbox{pink}{highlighed red}) and misses \textbf{keys facts} that the multi-agent model successfully captures (\textbf{bolded}).}
    \label{tab:summ3}
\end{table*}

\noindent\textbf{Independent vs. Communicating Agents} When trained on multiple agents with no communication (\textbf{m4}), the performance of our DCA models is similar to the single agent baselines (\textbf{m1}) and (\textbf{m3}). With communication, the biggest jump in ROUGE is seen on the CNN/DailyMail data, indicating that the encoders can better identify the key facts in the input, thereby avoiding unnecessary details. 

\noindent\textbf{Contextual Agent Attention (\textit{caa})} Compared to the model with no \emph{contextualized agent attention} (\textbf{m5}), the (\textbf{m6}) model yields better ROUGE scores. 
The stability provided by the \textit{caa} helps the decoder avoid frequent switches between agents that would dilute the topical signal captured by each encoder.  

\noindent\textbf{Repetition Penalty} 
As neurally generated summaries can be redundant, we introduced the semantic cohesion penalty and incremental rewards for RL to generate semantically diverse summaries. 
Our baseline model optimized together with SEM loss \textbf{(m2)} improves on all ROUGE scores over the baseline \textbf{(m1)}.   
Similarly, our model trained with reinforcement learning uses sentence based intermediate rewards, which also improves ROUGE scores across both datasets. 

\subsection{Human Evaluations}
\label{qualitativeanalysis}
We perform human evaluations to establish that our model's ROUGE improvements are correlated with human judgments. We measure the communicative multi-agent network with contextual agent attention in comparison to a single-agent network with no communication. 
We use the following as evaluation criteria for generated summaries: (1) \textit{non-redundancy}, fewer of the same ideas are repeated, (2) \textit{coherence}, ideas are expressed clearly; (3) \textit{focus}, the main ideas of the document are shared while avoiding superfluous details, and (4) \textit{overall}, the summary effectively communicates the article's content. 
The focus and non-redundancy dimensions help quantify the impact of multi-agent communication in our model, while coherence helps to evaluate the impact of the reward based learning and repetition penalty of the proposed models.
\begin{table}[t]
	\small
    \centering
\begin{tabular}{lccc|cc}
\hline
& \multicolumn{3}{c} {Head-to-Head} & \multicolumn{2}{c} {Score Based}\\
\cline{2-6}
Criteria & SA & MA & = & SA & MA \\ \hline
\hline
 non-redundancy &  68 & \textbf{159} & 73 & 4.384 & \textbf{4.428} \\
 coherence      &  89 & \textbf{173} & 38 & 3.686 & \textbf{3.754}\\
 focus          &  83 & \textbf{181} & 36 & 3.694 & \textbf{3.884$^{*}$}\\
\hline
 overall        &  102 & \textbf{158} & 40 & 3.558 & \textbf{3.682$^{*}$} \\
\hline
\end{tabular}
\vskip -0.1in
\caption{Head-to-Head and score-based comparison of human evaluations on random subset of CNN/DM dataset. SA=single, MA=multi-agent. $^*$ indicates statistical significance at $p < 0.001$ for focus and $p < 0.03$ for the overall.}
\label{human}
\end{table}

\noindent\textbf{Evaluation Procedure} 
We randomly selected 100 samples from the CNN/DailyMail test set and use workers from Amazon Mechanical Turk as judges to evaluate them on the four criteria defined above. Judges are shown the original document, the ground truth summary, and two model summaries and are asked to evaluate each summary on the four criteria using a Likert scale from 1 (worst) to 5 (best). The ground truth and model summaries are presented to the judges in random order. Each summary is rated by 5 judges and the results are averaged across all examples and judges.

We also performed a head-to-head evaluation (more common in DUC style evaluations) and randomly show two model generated summaries. We ask the human annotators to rate each summary on the same metrics as before without seeing the source document or ground truth summaries. 

\noindent\textbf{Results} Human evaluators significantly prefer summaries generated by the communicating encoders. In the rating task, evaluators preferred the multi-agent summaries to the single-agent cases for all metrics.  
In the head-to-head evaluation, humans consistently preferred the DCA summaries to those generated by a single agent. In both the head-to-head and the rating evaluation, the largest improvement for the DCA model was on the \textit{focus} question, indicating that the model learns to generate summaries with more pertinent details by capturing salient information from later portions of the document. 

\subsection{Communication improves focus}
To investigate how much the multi-agent models discover salient concepts in comparison to single agent models, we analyze ROUGE-L scores based on the average attention received by each agent. We compute the average attention received by each agent per decoding time step for every generated summary in the CNN/Daily Mail test corpus, bin the document-summary pairs by the attention received by each agent, and average the ROUGE-L scores for the summaries in each bin.

\begin{figure}[t]
\begin{center} 
\adjustbox{trim={0.04\width} {0.04\height} {0.04\width} {0.04\height},clip}%
{
\includegraphics[width=0.5\textwidth]{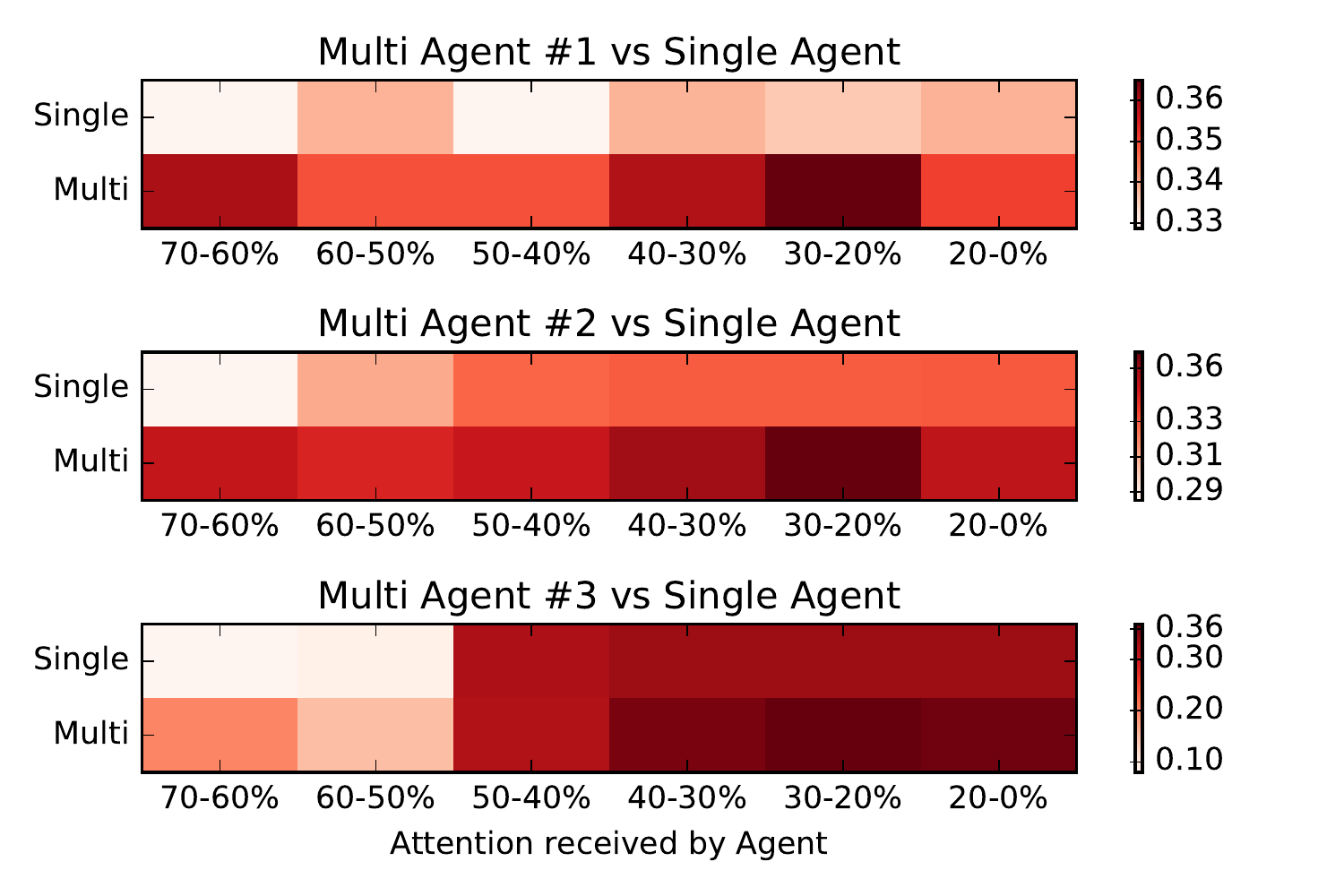}
}
\end{center} 
\vskip -0.18in
\caption{The average ROUGE-L scores for summaries that are binned by each agent's average attention when generating the summary (see Section~\ref{qualitativeanalysis}). When the agents contribute equally to the summary, the ROUGE-L score increases.} 
\label{agentattentiondetail}
\end{figure}

Figure~\ref{agentattentiondetail} outlines two interesting results. First, summaries generated with a more distributed attention over the agents yield higher ROUGE-L scores, indicating that attending to multiple areas of the document allows the discovery of salient concepts in the later sections of the text. 
Second, if we use the same bins and generate summaries for the documents in each bin using the single-agent model, the average ROUGE-L scores for the single-agent summaries are lower than for the corresponding multi-agent summaries, indicating that even in cases where one agent dominates the attention, communication between agents allows the model to generate more focused summaries.

Qualitatively, we see this effect in Table~\ref{tab:summ3}, where we compare the human generated summaries against our best single agent model (\textbf{m3}) and our best multi-agent model (\textbf{m7}). Model (\textbf{m3}) generates good summaries but does not capture all the facts in the human summary, while (\textbf{m7}) is able to include all the facts with few extra details, generating more relevant and diverse summaries. 

\section{Related Work} 
\label{sec:model}

Several recent works investigate attention mechanisms for encoder-decoder models to sharpen the context that the decoder should focus on within the input encoding \cite{att1,att2,bahdanau2014neural}. For example, \citet{att1} proposes {global} and {local} attention networks for machine translation, while others investigate hierarchical attention networks for document classification \cite{att3}, sentiment classification \cite{att4}, and dialog response selection \cite{att5}.

Attention mechanisms have shown to be crucial for summarization as well \cite{deepsummrush,efficientsum,summs2s}, and pointer networks \cite{pointer}, in particular, help address redundancy and saliency in generated summaries \cite{Lapata,summpoinernet,rlsummsocher,controllable}. While we share the same motivation as these works, 
our work uniquely presents an approach based on CommNet, the deep communicating agent framework \cite{commnet}.
Compared to prior multi-agent works on logic puzzles \cite{multiagent1}, language learning \cite{multiagent2,multiagent3} and starcraft games \cite{starcraft}, 
we present the first study in using this framework for long text generation. 

Finally, our model is related to prior works that address repetitions in generating long text.  
\citet{summpoinernet} introduce a post-trained coverage network 
to penalize repeated attentions over the same regions in the input, 
while
\citet{rlsummsocher} use intra-decoder attention to punish generating the same words.
In contrast, we propose 
a new semantic coherence loss and intermediate sentence-based rewards for reinforcement learning  
to discourage semantically similar generations (\S\ref{ssec:rewards}).

\section{Conclusions}
\label{sec:conclusion}
We investigated the problem of encoding long text 
to generate abstractive summaries and demonstrated that the use of 
deep communicating agents can improve summarization by both automatic and manual evaluation. Analysis demonstrates that this improvement is due to the improved ability of covering all and only salient concepts and maintaining semantic coherence in summaries.

\section*{Acknowledgements}
This research was supported in part by NSF (IIS-1524371), and DARPA under the CwC program through the ARO (W911NF-15-1-0543).
\bibliography{naaclhlt2018.bib}
\bibliographystyle{acl_natbib}
\newpage
\appendix
\section{Supplementary Material}
\label{appendix}
\begin{table}[h]
\small
\centering
\begin{tabular}{|l|c|c|}
 \hline
 \textbf{Stats} & \textbf{CNN/DM} & \textbf{NYT} \\
 \hline
 Avg. \# tokens document &781  & 549 \\
 Avg. \# tokens summary & 56 & 40 \\
 Total \# train doc-summ. pair & 287,229 & 589,284\\
 Total \# validation doc-summ. pair & 13,368 & 32,736\\
 Total \# test doc-summ. pair & 11,490 & 32,739\\
 Input token length & 400/800 & 800\\
 Output token length & 100 & 100 \\
 \hline
 (2-agent) Input token length / agent & 375 & 400\\
 (3-agent) Input token length / agent & 250 & 200\\
 (5-agent) Input token length / agent & 150 & 160\\
 \hline
\end{tabular}
\caption{Summary statistics of CNN/DailyMail (DM) and New York Times (NYT) Datasets.}
\label{tab:stats}
\end{table}
\subsection{Datasets}
\label{app:datasets}
\paragraph{CNN/DailyMail:}
CNN/DailyMail dataset \cite{summs2s,teachingmachines} is a collection of online news articles along with multi-sentence summaries.
We use the same data splits as in \citet{summs2s}. While earlier work anonymized entities by replacing each named entity with a unique identifier (e.g., \textit{Dominican Republic}$\rightarrow$\texttt{entity15}), we opted for non-anonymized version.

\paragraph{New York Times (NYT):}
Although this dataset has mainly been used to train extractive summarization systems \cite{nyt1,nyt2,nyt3,extractivesum}, it has recently been used for the abstractive summarization task \cite{rlsummsocher}. 
NYT dataset \cite{nytdataset} is a collection of articles published between 1996 and 2007. 
We use the scripts provided in \citet{nyt3} to extract and pre-process the NYT dataset with some modifications in order to replicate the pre-processing steps presented in \citet{rlsummsocher}.
Similar to \cite{rlsummsocher}, we sorted the documents by their publication date in chronological order and used the first 90\% for training, the next 5\% for validation and last 5\% for testing. They also use pointer supervision by replacing all named entities in the abstract if the type is "PERSON", "LOCATION", "ORGANIZATION" or "MISC" using the Stanford named entity recognizer \cite{corenlp}. By contrast, 
we did not anonymize the NYT dataset to reduce pre-processing.  

\subsection{Training Details}
\label{app:training}
We train our models on an NVIDIA P100 GPU machine.
We set the hidden state size of the encoders and decoders to 128.
For both datasets, we limit the input and output vocabulary size to the 50,000 most frequent tokens in the training set. 
We initialize word embeddings with 200-d GloVe vectors \cite{pennington2014glove} and fine-tune them during training.
We train using Adam with a learning rate of 0.001 for the MLE models and $10^{-5}$ for the MLE+RL models. 
We tune the $gamma$ hyper-parameter in the mixed loss by iterating $\gamma$=\{0.95, 0.97, 0.99\}. In almost all DCA models, the 0.97 value yielded the best gains.
We train our models for $~$200,000 iterations. 
which took 4-5 days for 2-3 agents and 5-6 days for 5 agents since it has more encoder parameters to tune.

To avoid repetition, we prevent the decoder from generating the same trigram more than once during test, following Paulus et.al.~(\citeyear{rlsummsocher}). In addition, for every predicted out-of-vocabulary token (UNK), we replace it with its most likely origin by choosing the source word \textit{w}
with the largest cascaded attention $w:=\text{arg}\max\limits_{a,i}l_{a,i}^t*g_a^t$ (Eq.~\eqref{attention2},~\eqref{aattention2}).

\section{Generated Summary Examples}
\label{app:generations}
This appendix provides example documents from the test set, with side-by-side comparisons of the human generated (golden) summaries
and the summaries produced by our models. \textbf{Baseline} is a \textbf{single-agent} model trained with MLE+RL loss, \textbf{(m3)} model in Table~\ref{tab:summ1}, while our best \textbf{multi-agent} model is optimized by mixed MLE+SEM+RL loss, the \textbf{(m7)} model in Table~\ref{tab:summ1}. 

\begin{itemize}
    \item \colorbox{pink}{red highlights} : indicate details that should not appear in the summary but the models generated them.
    \item \textcolor{red}{red} : indicates factual errors in the summary.
    \item \colorbox{lime}{green highlights} : indicate key facts in the human (gold) summary that only one of the models manage to capture.
\end{itemize}

\begin{table*}[t]
    \centering
\begin{center}
    \begin{tabular}{|P{1.5cm}@{}| p{13.8cm}|}
    \hline
    \small \textbf{Document} &model abbey clancy is helping to target breast cancer ,  by striking a sultry pose in a new charity campaign .   the winner of 2013 's strictly come dancing joins singer foxes ,  25 ,  victoria 's secret angel lily donaldson ,  28 ,  and model alice dellal ,  27 ,  in the new series of pictures by photographer simon emmett for fashion targets breast cancer .   clancy ,  29 ,  looks chic as she shows off her famous legs ,  wearing just a plain white shirt .   abbey clancy leads the glamour as she joins forces with her famous friends to target breast cancer ,  by striking a sultry pose in a new charity campaign  the model ,  who is mother to four - year - old daughter sophia with footballer husband peter crouch ,  said: ' as a mum ,  it makes me proud to be part of a campaign that funds vital work towards ensuring the next generation of young women do not have be afraid of a diagnosis of breast cancer .  ' i'm wearing my support ,  and i want everyone across the uk to do the same and get behind this campaign .  ' holding onto heaven singer foxes looks foxy in cropped stripy top and jeans .   abbey says she is proud to be part of a campaign that funds vital work towards ensuring the next generation of young women do not have be afraid of a diagnosis of breast cancer  victoria 's secret angel lily donaldson ,  who has been in the industry for years ,  also adds some glamour to the charity campaign  holding onto heaven singer foxes dons a stripy top and jeans for the campaign she says she 's ' honoured ' to be a part of  she said: ' i'm so honoured to be taking part in this year 's fashion targets breast cancer ,  and becoming part of the campaign 's awesome heritage .  ' fashion is a huge part of my life ,  and if by taking part i can inspire women to wear their support ,  join the fight and take on breast cancer head on ,  then that will be something to be really proud of .  ' now in its 19th year ,  the campaign has so far raised 13 . 5m for breakthrough breast cancer 's research funding .   this year the range of clothes and accessories have been produced in conjunction with high street partners m\&s ,  river island ,  warehouse ,  topshop ,  laura ashley ,  debenhams ,  superga ,  baukjen and the cambridge satchel company .   they can be viewed online at www . fashiontargetsbreastcancer . org . uk/lookbook  the campaign ,  which also stars alice dellal ,  has so far raised 13 . 5m for breakthrough breast cancer 's research funding 

    \\ \hline
    \small  \textbf{Human (Gold)}  & models abbey and lily are joined by alice dellal and singer foxes . the women are pictured ' wearing ' their support . abbey , 29 , says she is proud to be part of a campaign that funds vital work . campaign has raised 13 . 5m for breakthrough breast cancer 's research .
 \\ \hline
  \small  \textbf{Single Agent Baseline} &  \colorbox{pink}{strictly come dancing joins singer foxes , 25 , victoria 's secret angel lily donaldson , 28} , and model alice dellal , 27 , in the new series of pictures by photographer simon emmett for fashion targets breast cancer . clancy , 29 , looks chic as she shows off her famous legs , wearing just a plain white shirt .

  \\\hline
  \small  \textbf{Multi Agent} &  abbey says she is proud to be part of a campaign that funds vital work towards ensuring the next generation of young women do not have been afraid of a diagnosis of breast cancer . the \colorbox{lime}{campaign has raised 13 . 5m for breakthrough breast cancer 's research }.

 \\ \hline
    \end{tabular}
\end{center}
\vspace{-0.2in}
\caption{In this example both single- and multi-agent models demonstrate extractive behaviors. However, each select sentences from different sections of the document. While the single model extracts the second and the third sentences, the multi-agent model successfully selects salient sentences from sentences that are further down in the document, specifically sentence 8 and 10. This can be attributed to the fact that agents can successfully encode salient aspects distributed in distant sections of the document. An interesting result is that even though the multi-agent model shows extractive behaviour in this example, it successfully selects the most salient sentences while the single agent model includes superfluous details.}
    \label{app:sum1}
\end{table*}

\begin{table*}[t]
    \centering
\begin{center}
    \begin{tabular}{|P{1.5cm}@{}| p{13.8cm}|}
    \hline
    \small \textbf{Document} & michelle pfeiffer is the latest hollywood star preparing to hit the small screen .   the oscar nominated star known for her roles in iconic films such as scarface , dangerous liaisons andthe age of innocence ,  has teamed up with katie couric to pitch a new television comedy about a morning news program .   also involved in the project ,  in which pfeiffer is attached to star ,  is diane english ,  the creator of murphy brown .   scroll down for video  michelle pfeiffer ( left ) is set to star in a new television comedy about a morning news program produced by katie couric ( right )  the series was created by diane english ( above with candice bergen ) ,  who was behind the show murphy brown ,  about a female news anchor  according to variety ,  pfeiffer 's role will be that of a morning news anchor ,  making it very similar to the real life role couric played as co - host of today for 15 years . \textbf{couric will serve as an executive producer and help ' ensure the series strikes realistic notes .  ' the creator behind the project ,  english ,  was previously the brains behind brown ,  the show starring candice bergen that centered around a female news anchor and ran for ten seasons ,  winning 18 emmys .   english would also serve as a writer ,  producer and showrunner on the program}.   the ladies are currently in talks with hbo ,  showtime ,  amc ,  netflix and amazon to pick up the program .   couric will serve as an executive producer ,  drawing on her experience as an anchor on today for 15 years  pfeiffer would be the one of the biggest stars yet to move to television ,  joining a group that now includes house of cards stars robin wright and kevin spacey ,  true detective leads matthew mcconaughey and woody harrelson ,  and even lady gaga ,  who recently announced she would be appearing on the next season of american horror story .   the actress has kept a low profile for the past 20 years since becoming a mother ,  only doing a handful of films in that time .   she most recently appeared alongside robert de niro in the mob comedy ' the family .  '

    \\ \hline
    \small  \textbf{Human (Gold)}  & michelle pfeiffer is set to star in a new television comedy about a morning news program . katie couric will serve as an executive producer , drawing on her experience as an anchor on today for 15 years . the series was created by diane english , who was behind the show murphy brown , about a female news anchor . the ladies are currently in talks with hbo , showtime , amc , netflix and amazon to pick up the program .
 \\ \hline
  \small  \textbf{Single Agent Baseline} & the oscar nominated star \colorbox{pink}{known for her roles in iconic filmssuch as scarface , dangerous} \colorbox{pink}{liaisons and the age of innocence} , has teamed up with katie couric to pitch a new television comedy about a morning news program . also involved in the project , in which pfeiffer is attached to star , is diane english , the creator of murphy brown .

  \\\hline
  \small  \textbf{Multi Agent} &  michelle pfeiffer is set to star in a new tv comedy about a morning news program . \colorbox{lime}{couric will} \colorbox{lime}{serve as an executive producer and showrunner on the project .} the series was created by diane english , the creator of murphy brown . pfeiffer is the one of the biggest stars .
 \\ \hline
    \end{tabular}
\end{center}
\vspace{-0.2in}
\caption{The baseline model generates non-coherent summary that references the main character ``Michelle Pfeiffer" in an ambiguous way towards the end of the generated summary. In contrast, the multi-agent model successfully captures the main character including the \textbf{key facts}. One interesting feature that the multi-agent model showcases is its simplification property, which accounts for its strength in \textbf{abstraction}. Specifically, it simplified the \textbf{bold} long sentence in the document starting with "\textit{couric will...} and only generated the salient words.}
    \label{app:sum2}
\end{table*}

\begin{table*}[t]
    \centering
\begin{center}
    \begin{tabular}{|P{1.5cm}@{}| p{13.8cm}|}
    \hline
    \small \textbf{Document} & everton manager roberto martinez was forced to defend another penalty fiasco at the club after ross barkley missed from the spot in their 1 - 0 win against burnley at goodison park .   the untried barkley inexplicably took the 10th minute kick  awarded for a foul by david jones on aaron lennon  rather than leighton baines ,  who has scored 15 penalties from 16 attempts in the premier league .   although there was no dispute between the team - mates this time ,  it brought back memories of everton 's match against west brom in january when kevin mirallas grabbed the ball from baines to take a penalty  -  and missed .   ross barkley steps up to take a 10th minute penalty despite the presence of leighton baines on the pitch  barkley 's effort is saved byburnley goalkeeper tom heaton at goodison park  martinez insisted barkley was within his rights to request penalty - taking duties on saturday .  ' if romelu lukaku had been on the pitch ,  he would have taken it .  otherwise ,  i am happy to have three or four players who can take penalties and let it depend on how they feel at that moment ,  ' argued the everton manager .   baines ( left )has scored 15 penalties from 16 attempts in the premier league ' ross showed incredible responsibility to take it .  i love seeing players take control of the big moments and leighton was happy to given him that responsibility .  ' barkley 's penalty was well - struck but wasn't put in the corner and burnley goalkeeper tom heaton dived to his right to save .   fortunately for the young england player ,  it didn't prove costly as mirallas went on to score the only goal of the game after 29 minutes .   everton boss roberto martinez issues instructions to his players during a break in play against burnley
    \\ \hline
    \small  \textbf{Human (Gold)}  & everton defeated burnley 1 - 0 at goodison park on saturday . kevin mirallas scored the only goal of the game in the 29th minute . ross barkley had earlier missed a 10th - minute penalty . leighton baines has scored 15 penalties from 16 attempts this season .
 \\ \hline
  \small  \textbf{Single Agent Baseline} & \colorbox{pink}{everton manager roberto martinez was forced to defend another penalty fiasco at the club} \colorbox{pink}{after ross barkley missed from the spot} in their 1 - 0 win against burnley at goodison park . the untried barkley inexplicably took the 10th minute kick awarded for a foul by david jones on aaron lennon rather than leighton baines , who has scored 15 penalties from 16 attempts in the premier league .
  \\\hline
  \small  \textbf{Multi Agent} &  everton beat burnley 1 - 0 at goodison park in the premier league . ross barkley steps up to take a 10th minute penalty but missed it . barkley has scored 15 penalties from 16 attempts in the pitch .
 \\ \hline
    \end{tabular}
\end{center}
\vspace{-0.2in}
\caption{The single agent model generates summary with superfluous details and the facts are not clearly expressed. Although it was able to capture the statistics of the player correctly (e.g., \textit{15 penalties}, \textit{16 attempts}), it still missed the player who scored the only goal in the game (i.e., \textit{kevin mirallas}). On the other hand multi-agent model was able to generate a concise summary with several key facts. However, similar to single agent model, it missed to capture the player who scored the only goal in the game. Interestingly, the document contains the word "\textit{defeated}' but the multi-agent model chose to use \textit{beat} instead, which does not exist in the original document.}
    \label{app:sum3}
\end{table*}

\end{document}